%% file: StackVis.tex
\documentclass[journal]{vgtc}                

\usepackage{silence}
\usepackage[absolute,overlay]{textpos}
\usepackage{pbox}
\usepackage{everypage}

\newcommand{\stamp}[1][© 2021 IEEE. This is the author's version of the article that has been published in IEEE Transactions on Visualization and Computer Graphics. The final version of this record is available at: \href{https://doi.org/10.1109/TVCG.2020.3030352}{\color{blue}10.1109/TVCG.2020.3030352}]{%
\begin{textblock*}{140mm}(37mm,270mm)
\centering%
\small%
\emph{#1}%
\end{textblock*}%
}

\AddEverypageHook{
  \stamp
}

\ifpdf
  \pdfoutput=1\relax                   
  \usepackage{graphicx}                
  \DeclareGraphicsExtensions{.pdf,.png,.jpg,.jpeg} 
\else
  \usepackage{graphicx}                
  \DeclareGraphicsExtensions{.eps}     
\fi%

\graphicspath{{figures/}{pictures/}{images/}{./}} 

\usepackage{microtype}                 
\PassOptionsToPackage{warn}{textcomp}  
\usepackage{textcomp}                  
\usepackage{mathptmx}                  
\usepackage{times}                     
\usepackage{cite}                      
\usepackage{tabu}                      
\usepackage{booktabs}                  

\usepackage[dvipsnames]{xcolor} 
\usepackage[caption=false]{subfig}

\onlineid{0}

\vgtccategory{Research}
\vgtcpapertype{application/design study}

\WarningsOn

\newcommand{\cit}[1]{``#1''}
\newcommand{\circled}[1]{\raisebox{.4pt}{\textcircled{\raisebox{-.0pt} {\tiny \bf  #1}}}}

\title{StackGenVis: Alignment of Data, Algorithms, and Models for Stacking Ensemble Learning Using Performance Metrics}

\author{Angelos Chatzimparmpas, \textit{Student Member, IEEE}, Rafael M.\ Martins, \textit{Member, IEEE Computer Society},\\%
Kostiantyn Kucher, \textit{Member, IEEE Computer Society}, and Andreas Kerren, \textit{Senior Member, IEEE}}

\authorfooter{
\item
Angelos Chatzimparmpas, Rafael M. Martins, Kostiantyn Kucher, and Andreas Kerren are with Linnaeus University, V{\"a}xj{\"o}, Sweden. 
\\E-mail: $\left\{\begin{varwidth}{12cm}\centering angelos.chatzimparmpas, rafael.martins,\\kostiantyn.kucher, andreas.kerren\end{varwidth}\right\}$@lnu.se} 

\shortauthortitle{Chatzimparmpas \MakeLowercase{\textit{et al.}}: StackGenVis}

\abstract{
  \input{0.Abstract}
}

\keywords{Stacking, stacked generalization, ensemble learning, visual analytics, visualization}

\teaser{
 \centering
 \includegraphics[width=\linewidth]{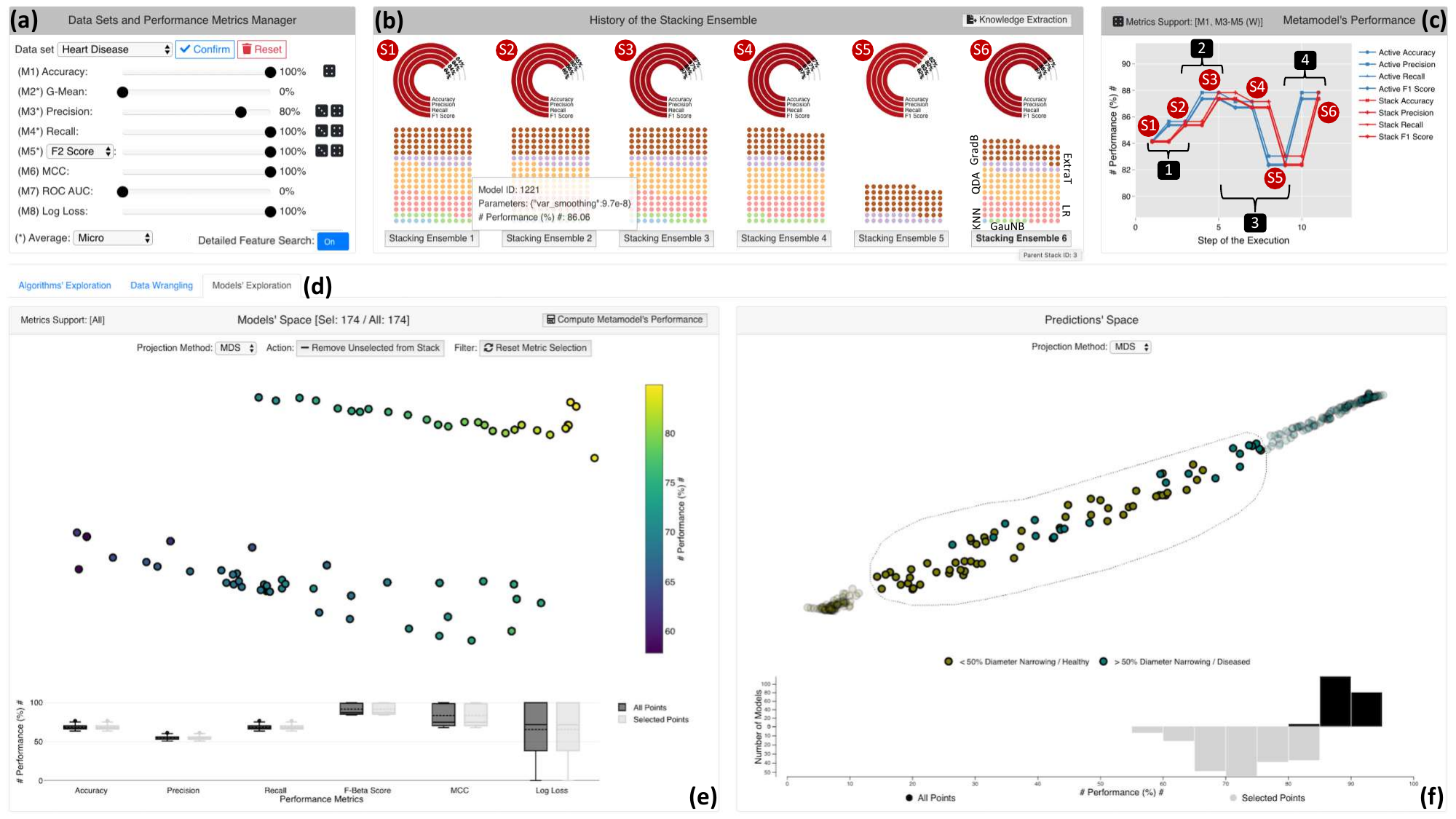}\vspace{-3mm}
 \caption{Constructing performant stacking ensembles from scratch with StackGenVis: (a) a panel for uploading data sets and choosing weights for performance metrics; (b) the history preservation panel with the composition and performance achieved by the user-built stored stacking ensembles; (c) the comparison of the metamodel's performance for both the active and stored stackings, based on four performance metrics (linked to view (a) with a dice glyph showing four); (d) the three exploration modes for the algorithms, data, and models; (e) the projection-based models' space visualization, which summarizes the results of all the selected performance metrics for all models; and (f) the predictions' space visual embedding, which arranges the data instances based on the collective outcome of the models in the current stored stack \circled{S6} (marked in bold typeface in (b)).}
	\label{fig:teaser}
}

\vgtcinsertpkg
\WarningsOn

\begin{document}


\stamp

\maketitle

\section{Introduction} \label{sec:intro}{%
  \input{1.Introduction}
}

\section{Related Work} \label{sec:relwork}{%
  \input{2.Related_Work}

\section{Design Goals and Analytical Tasks} \label{sec:stacking}
  \input{3.Stacking}

\section{StackGenVis: System Overview and Application} \label{sec:stackgenvis} {%
  \input{4.System_Overview}
 
\section{Use Case} \label{sec:case} {%
  \input{5.Use_Case}

\section{Evaluation and Future Work} \label{sec:disc} {%
  \input{6.Evaluation_Discussion}

\section{Conclusion} \label{sec:con} {%
  \input{7.Conclusion}

\bibliographystyle{abbrv-doi}

\bibliography{references} 

\end{document}

%% file: 0.Abstract.tex
In machine learning (ML), ensemble methods---such as bagging, boosting, and stacking---are widely-established approaches that regularly achieve top-notch predictive performance. Stacking (also called \cit{stacked generalization}) is an ensemble method that combines heterogeneous base models, arranged in at least one layer, and then employs another metamodel to summarize the predictions of those models. Although it may be a highly-effective approach for increasing the predictive performance of ML, generating a stack of models from scratch can be a cumbersome trial-and-error process. This challenge stems from the enormous space of available solutions, with different sets of data instances and features that could be used for training, several algorithms to choose from, and instantiations of these algorithms using diverse parameters (i.e., models) that perform differently according to various metrics. In this work, we present a knowledge generation model, which supports ensemble learning with the use of visualization, and a visual analytics system for stacked generalization. Our system, StackGenVis, assists users in dynamically adapting performance metrics, managing data instances, selecting the most important features for a given data set, choosing a set of top-performant and diverse algorithms, and measuring the predictive performance. In consequence, our proposed tool helps users to decide between distinct models and to reduce the complexity of the resulting stack by removing overpromising and underperforming models. The applicability and effectiveness of StackGenVis are demonstrated with two use cases: a real-world healthcare data set and a collection of data related to sentiment/stance detection in texts. Finally, the tool has been evaluated through interviews with three ML experts.

%% file: 1.Introduction.tex
Stacking methods (or \emph{stacked generalizations}) refer to a group of ensemble learning methods~\cite{Sagi2018Ensemble} where several base models are trained and combined into a metamodel with improved predictive power~\cite{Wolpert1992Stacked}. In particular, stacked generalization can reduce the bias and decrease the generalization error when compared to the use of single learning algorithms. To accomplish that, stacking enables the blending of different and heterogeneous \emph{algorithms} and their instantiations with particular parameters, i.e., \emph{models}. Other types of ensemble methods are \emph{bagging techniques}, such as random forests (RF)~\cite{Breiman2001Random}, and \emph{boosting techniques}, such as adaptive boosting (AdaB)~\cite{Freund1999A} or gradient boosting (GradB)~\cite{Chen2016XGBoost,Ke2017LightGBM}. A major difference between these ensemble methods is that stacking can use both bagging and boosting techniques in combination with simpler algorithms, stacked in different layers. 
It uses a \emph{meta-learner} to aggregate the predictions of the last layer and obtain the best performance, which is absent in the other ensemble methods.

In numerous Kaggle competitions~\cite{Kaggle2015}, stacking ensembles led to award-winning results. But, when studying such ensembles, it is very hard to understand \emph{why} specific instances, features, algorithms, and models were selected instead of others. Indeed, one of the major challenges in stacking is to select the best combinations of algorithms and models when designing a stacking ensemble from scratch. This issue may keep machine learning (ML) practitioners and experts away from working with complex stacking ensemble methods, even though they could arguably reach very high-performance results. 
One question that arises from the work by Naimi and Balzer~\cite{Naimi2018Stacked} is: \textbf{(RQ1)} how to build a stacking ensemble for a given problem with a focus on avoiding such trial and error methods, and/or increasing the overall efficiency?

In spite of this challenge of hardly understanding why a specific configuration works~\cite{Ting1997Stacked}, predicting the relation of supply-demand~\cite{Tugay2017Demand} and anomaly/bug reports~\cite{Jonsson2012Towards,Jonsson2016Automated} are areas where stacking has been used successfully. Compelling accuracy results~\cite{Sospedra2006Combining} were also observed for text data, where stacking is better than alternative techniques such as voting ensembles~\cite{Sigletos2005Combining}. Above all, mixtures of stacked models have been deployed to increase the performance of results in medicine~\cite{Jyothi2019A,Ma2018Ensemble,Nagi2013Classification}. 
In the case of healthcare-related problems, however, the difficulties of stacking lead to an even worse situation, because interpretability, fairness in decisions, and trustworthiness of ML models are very critical in the medical field~\cite{Cohen2016STARD}. 
The recent survey by Sagi and Rokach~\cite{Sagi2018Ensemble} lists the users' ability to understand how to tune the models as one of the important factors for selecting the appropriate ensemble learning method, too.
Thus, another open question is: \textbf{(RQ2)} how to monitor and control the complete process of training stacking ensembles, while preserving confidence and trust in their predictive results?

Performance metrics, such as precision or f1-score, are typically adopted to validate if the ML results meet the expectations of the experts and the domain~\cite{Ferri2009An,Pereira2017A,Sokolova2009Performance,Tharwat2018Classification}.
Multiple metrics are important to avoid the dangers of using single metrics, such as accuracy~\cite{McNee2006Being,Sturm2013Classification}, for every data set.
However, comparison and selection between multiple performance indicators is not trivial, even for widely used metrics~\cite{Davis2006The,Saito2015The}; alternatives such as Matthews correlation coefficient (MCC) might be more informative for some problems~\cite{Chicco2020The}.
Further open challenges of using advanced metrics are described in the literature~\cite{Lobo2008AUC,Powers2011Evaluation}. 
This leads to one further question: \textbf{(RQ3)} what performance/validation metrics fit better to a specific data set, and how can they be combined?

Stacking ensemble learning inspired us to focus on each of the three aforementioned questions that represents an open research challenge. In this paper, we present a knowledge generation model for ensemble learning with the use of visualization (derived from Sacha et al.~\cite{Sacha2014Knowledge}), 
and instantiate this model as our visual analytics (VA) system for stacked generalization. 
Our system, called StackGenVis (see \autoref{fig:teaser}), tries to address the three questions described above by supporting the exploratory combination of 11 different ML algorithms and 3,106 individual models using 8 performance metrics with various modes (see the details in \autoref{sec:stackgenvis}).
To address those three open research challenges (\textbf{RQ1--RQ3}), StackGenVis supports the following \textbf{workflow}: (i) the selection of appropriate validation metrics, (ii) the exploration of algorithms, (iii) the data wrangling, (iv) the exploration of models, and (v) an overarching phase, where the resulting stack is traced and the performance of the stored stack is compared to the current active metamodel.
In summary, our contributions consist of the following:

\begin{itemize}
\item the composition of a knowledge generation model (KGM) specifically adapted for ensemble learning with the use of VA;
\item the implementation of a VA system, called StackGenVis, that follows the KGM mentioned above, consists of novel views that treat models and predictions as high-dimensional vectors, and supports the visual exploration of the most performant and most diverse models for the creation of stacking ensembles;
\item the applicability of our proposed system with two use cases, using real-world data, that confirm the effectiveness of utilizing multiple validation metrics and comparing stacking ensembles; and
\item the discussion of the followed methodology and the outcomes of several expert interviews that indicate encouraging results.
\end{itemize}      

\noindent The rest of this paper is organized as follows. In the next section, we discuss the literature related to visualization of ensemble learning.
Afterwards, we describe the knowledge generation model for ensemble learning with VA, design goals, and analytical tasks for attaching VA to ensemble learning.
\autoref{sec:stackgenvis} presents the functionalities of the tool and, at the same time, describes the first use case with the goal of improving another stacking ensemble's results for healthcare data.
Next, we demonstrate the applicability and usefulness of StackGenVis with our own real-world data set focusing on sentiment/stance in texts. Thereafter in~\autoref{sec:disc}, we discuss the feedback our VA system received during the conducted expert interviews by summarizing the opinions of the experts and the limitations that lead to possible future work for our approach. Finally,~\autoref{sec:con} concludes our paper.

%% file: 2.Related_Work.tex
\begin{figure*}[tb]
\centering
\includegraphics[width=\linewidth]{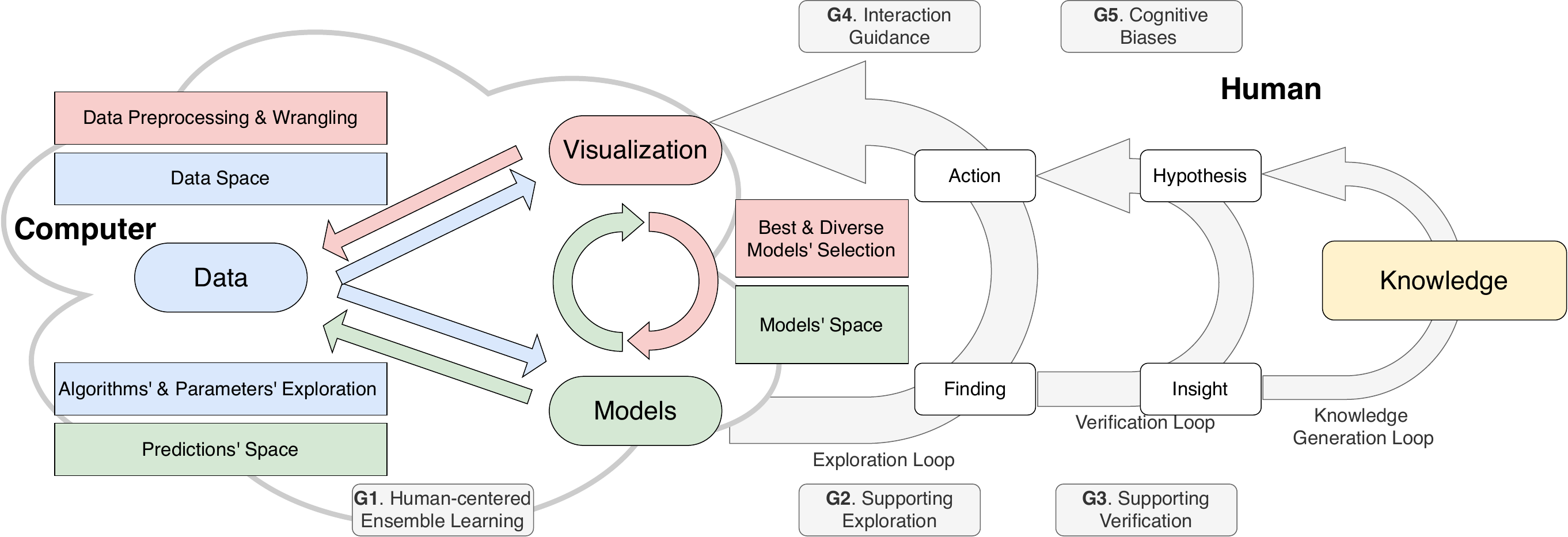}
\caption{Knowledge generation model for ensemble learning with VA derived from the model by Sacha et al.~\cite{Sacha2014Knowledge}. On the left, it illustrates how a VA system can enable the exploration of the data and the models with the use of visualization. On the right, a number of design goals assist the human in the exploration, verification, and knowledge generation for ensemble learning.}
\label{fig:framework}
\end{figure*}

Visualization systems have been developed for the exploration of diverse aspects of bagging, boosting, and further strategies such as  ``bucket of models''. 
Stacking, however, has so far not received comparable attention by the InfoVis/VA communities: actually, we have not found any literature describing the construction and improvement of stacking ensemble learning with the use of VA. 
In this section, we briefly discuss previous works on bagging, boosting, and buckets of models, and highlight their differences with StackGenVis in order to substantiate the novelty of our approach.

\subsection{Bagging and Boosting} \label{sec:bagging}

EnsembleLens~\cite{Xu2019EnsembleLens} is a VA system that focuses on the identification of the best combination of models by visualizing their correlation. Specific feature subsets are chosen to train each algorithm---a technique known as \emph{feature bagging}. Then, the results are combined and ranked based on the performance outcomes for anomalous cases.
In contrast, our work is not limited to the anomaly detection task, and it focuses on construction of better-performing ensembles by combining multiple algorithms and using appropriate performance metrics.

BEAMES~\cite{Das2019BEAMES} focuses on regression tasks, and it includes four learning algorithms and a model sampling technique.
The output of the system is a ranking that helps the user to decide on a model. 
In our approach, a metamodel automatically chooses well-performing models. 
BEAMES includes three performance metrics: (a) residual error, (b) mean squared error, and (c) $r$-squared error, which are specialized for regression problems. Interestingly, the authors suggest as future work that \cit{there are open research questions about how best to compare multiple models directly}. In our system, we address this challenge with the exploration of a finite space of solutions, employing 11 algorithms (that can be further expanded). 
Exploration of feature importance and instances in BEAMES involves a standard table representation and recommendation of best-performing models for specific instances. 
In StackGenVis, we present different ways for direct manipulation of instances, highlighting hard-to-classify instances. 
Three separate techniques are incorporated for feature selection, visualized by an aggregated table heatmap view.

iForest~\cite{Zhao2019iForest} is a VA system that uses dimensionality reduction (DR) to summarize the predictions of each instance; other views explain  decision paths of an RF. 
It also highlights the relationship between the features of the data set and the prediction outcomes.
For a specific instance, a new DR projection can be used to show which models performed well or not, and why. 
The goals and challenges addressed by iForest are different than ours: it strives to open the black box of a specific algorithm, while StackGenVis uses a parallel and model-agnostic strategy accompanied by high-level monitoring of the process.
Additionally, we utilize multiple validation metrics simultaneously to explore diverse models, instead of relying on decision trees only.

Similarly to iForest, BoostVis~\cite{Liu2018Visual} also uses DR and other views to compare and improve ML algorithms such as XGBoost~\cite{Chen2016XGBoost} or LightGBM~\cite{Ke2017LightGBM}. The goal is to diagnose and debug the training process of underperforming trees, which are visualized with trajectories in a DR projection. 
Our work, in contrast, focuses on the appropriate selection of models to enhance---as much as possible---the prediction power of a stacking ensemble. 
Moreover, we use three alternative techniques to rank the most important features for several hundreds or thousands of models, and we use multiple performance metrics, with user-defined weightings, to characterize the results.

Schneider et al.~\cite{Schneider2018Integrating} employed both bagging and boosting ensembles in an effort to combine the data and model space. The authors applied scatterplots and DR projections for the visualization of the data space, with the goal to add, delete, or replace models from the ensemble model space.
Pairs of validation metrics allow the user to select the best models (sorted by performance or similarity). A selection results in an update of the data space.
Our approach of aligning the data and model spaces is influenced by this work, but we improved the process by aggregating the alternative performance metric results on top of the projections.
Furthermore, we allow the users to define specific weights for each metric and focus on the models that perform well for both the entire data space and specific instances. 
Finally, StackGenVis does not support direct manipulation of model ensembles~\cite{Schneider2018Integrating}, as it focuses on exploration of a large solution space before narrowing down to specific well-performing and diverse models.

\subsection{Buckets of Models} \label{sec:buckets}

In a bucket of models, the best model for a specific problem is automatically chosen from a set of available options. This strategy is conceptually different to the ideas of bagging, boosting, and stacking, but still related to ensemble learning. 
Chen et al.~\cite{Chen2019LDA} utilize a bucket of latent Dirichlet allocation (LDA) models for combining topics based on criteria such as distinctiveness and coverage of the set of actions performed.
Pie charts on top of projections show probability distributions of action classes. Although this work is not similar to StackGenVis in general, we use a gradient color scale to map the performance of each model in the projected space. 
EnsembleMatrix~\cite{Talbot2009EnsembleMatrix} linearly fuses multiple models with the help of a confusion matrix representation, while supporting comparison and contrasting for model exploration.  
In our VA system, the user can explore how models perform on each class of the data set, and the performance metrics are instilled into a combined user-driven value. Manifold~\cite{Zhang2019Manifold} generates pairs of models and compares them over all classes of a data set, including feature selection. We adopt a similar approach, but instead of comparing a large number of models in a pairwise manner, we aggregate their overall and per-class performance. Then, the user can compare a set of models against the average of all the models before deciding which ones to use.  

There is also a group of works that focuses specifically on regression problems~\cite{Muhlbacher2013APartition,Sehgal2018Visual,Zhao2014LoVis}. 
For instance, the more recent tool iFuseML~\cite{Sehgal2018Visual} operates with prediction errors in order to present ensemble models with more accurate predictions to the users. The comparison of models is very different in our approach: we use preliminary results from performance metrics in order to select the appropriate models that will boost the final stack performance.

%% file: 3.Stacking.tex
In this section, 
we explain the main design goals that base the development of StackGenVis, together with a knowledge generation model (KGM) for ensemble learning~(\autoref{fig:framework}).
Then, we describe the analytical tasks that StackGenVis (and any other VA system) should tackle in order to support the presented KGM with regard to stacking methods. 

\subsection{Design Goals: Visual Analytics to Support Ensemble Learning}

In the following, we define five design goals (\textbf{G1--G5}) built on top of the knowledge generation model for VA proposed by Sacha et al.~\cite{Sacha2014Knowledge}.
This original model has two core pillars: the \emph{computer} (\autoref{fig:framework}, left) and the \emph{human} (\autoref{fig:framework}, right). 
On the computer side, the VA system comprises \emph{data}, \emph{visualization(s)}, and \emph{model(s)}. 
The human side depicts the knowledge generation process, comprising the loops for \emph{exploration},  \emph{verification}, and \emph{knowledge generation}. 

Our design goals focus on the knowledge generation in ensemble learning with the use of VA, originating from the analysis of the related work in~\autoref{sec:relwork}, our own experiences when developing VA tools for ML (e.g., t-viSNE~\cite{Chatzimparmpas2020t}), and recently conducted literature reviews~\cite{Chatzimparmpas2020A,Chatzimparmpas2020The}. We slightly extended the original knowledge generation model for VA~\cite{Sacha2014Knowledge} to make a better fit for supporting ensemble learning with VA (cf. the description of design goal \textbf{G1}) and then aligned our design goals to the different model parts, see the gray boxes in \autoref{fig:framework}.

\textbf{G1: Incorporate human-centered approaches for controlling ensemble learning.} 
For our first design goal, we modified the original knowledge generation model for VA~\cite{Sacha2014Knowledge} by adding components specifically related to ensemble methods~\cite{Sagi2018Ensemble}. 
Ensemble learning can be controlled in different ways. Starting from the data, visualization can be used to explore the \emph{data space} (\autoref{fig:framework}, upper blue arrow)~\cite{Schneider2018Integrating}. This offers new possibilities for direct manipulation of both instances and features. Visualization also enhances the interaction with \emph{data preparation} (\autoref{fig:framework}, upper red arrow)~\cite{Liu2018Visual}. Data preprocessing and wrangling benefits from feedback provided by a VA system, for example, in the form of validation metrics that increase the per-model performance of several heterogeneous ML models used in ensemble learning. 
Next, VA is useful for the exploration and final selection of different \emph{algorithms} that have numerous \emph{parameters} leading to well-performing models  (\autoref{fig:framework}, lower blue arrow)~\cite{Das2019BEAMES}. These models produce predictions that can be stored again as new metadata. If visualized, this \emph{predictions' space} can be manipulated accordingly for raising the overall predictive performance (\autoref{fig:framework}, lower green arrow). 
The process of ensemble learning generates a \emph{solution space of models} (\autoref{fig:framework}, curved green arrow)~\cite{Schneider2018Integrating} and more investigations can be done to choose between the \emph{best and most diverse models} of an ensemble (\autoref{fig:framework}, curved red arrow). The careful design, choice, and arrangement of these aspects and the balance between human-centered vs. automated approaches are essential concepts when developing a VA system~\cite{Shneiderman2020Human}. Moreover, the different perspectives of analysts working on a problem can push toward more efficient and effective solutions or receiving results in a shorter amount of time. Synchronous and asynchronous collaboration can empower visualizations dedicated to particular tasks~\cite{Isenberg2011Collaborative}. Building ensembles from scratch by using various ML algorithms might require expert collaboration and intervention, especially when those experts are specialized on individual algorithms. If a VA system supports asynchronous and/or synchronous communication, an individual expert can share his/her knowledge with the others, which could lead to a more desirable outcome.

\textbf{G2: Support exploration.} VA systems enable users to reach crucial \emph{findings} and to take \emph{actions} according to them. This iterative process requires a human-in-the-loop who can thus explore the data and the model through the interactive visualization~\cite{Brehmer2013A}. 
As the solution space for ensemble learning is more confusing compared to single ML techniques, keeping track of the history of events and providing provenance for exploring and backtracking of alternative paths is necessary to reach this goal.
Furthermore, provenance in VA for ensemble learning increases the interpretability and explainability caused by the complex nature of the method.
Although provenance in VA systems has been in the research focus during the past years~\cite{Oliveira2017A,Ragan2016Characterizing}, the work on utilizing analytic provenance~\cite{Xu2015Provenance} is still limited.

\textbf{G3: Support verification.}
According to the \emph{insights} gained from the exploration process, users are able to formulate new \emph{hypotheses} that can be efficiently tested with the help of interactive visualization.
This goal is valuable especially for ensemble methods, which are harder to train and verify than individual ML models. Annotations within a visualization are used to share insights between analysts or to save information for later use. In storytelling, for example, the annotation is considered as a key element~\cite{Tong2018Storytelling}.
Keeping notes linked to particular views of a VA system for ensemble learning could be essential for remembering key findings and core actions for reaching good performance results.

\textbf{G4: Facilitate human interaction and offer guidance.} During development of any VA tool, it is key to decide on concrete visual representations and interaction technologies between multiple coordinated views. It is not uncommon to find gaps between visualization design guidelines and their applicability in implemented tools~\cite{Moritz2019Formalizing}, and providing guidance in the complex human-machine analytical process is another challenge~\cite{Collins2018Guidance}.
Many different facets are involved in VA for ensemble learning, ranging from diverse ML models and data sets to performance metrics.
From a visualization perspective, this heterogeneity leads to multiple views.
A careful visual design of the linked views that facilitate human interaction and sophisticated VA systems that guide the user to important aspects will help to disentangle the visual complexity and, in consequence, the cognitive load of the user.

\textbf{G5: Reveal and reduce cognitive biases.} Visualizations should be carefully chosen in order to reduce cognitive biases. Cognitive bias is, in simple terms, a human judgment that drifts away from the actual information that should be conveyed by a visualization, i.e., it \cit{involves a deviation from reality that is predictable and relatively consistent across people}~\cite{Dimara2020A}. 
The use of visualization for ensemble learning could possibly introduce further biases to the already blurry situation based on the different ML models involved. Thus, the thorough selection of both interaction techniques and visual representations that highlight and potentially overcome any cognitive biases is a major step toward realizing this design goal.

\subsection{Analytical Tasks for Stacking} 

To fulfill our design goals specifically in the context of stacking ensemble learning, we have derived five high-level analytical tasks that should be solved by our VA system described in \autoref{sec:stackgenvis}.

\textbf{T1: Search the solution space for the most suitable algorithms, data, and models for the task.}
Some of the major challenges of stacking are the choice of the most suitable algorithms and models, the data processing necessary for the selected models, further improvements for the models, and reduction of the complexity of the stack (\textbf{G1}).
This workflow should be assisted by guidance at different levels, including the selection of proper performance metrics for the particular problem and the comparison of results against the current stack.

\textbf{T2: Explore the history with all basic actions of the stacking ensemble preserved.}
There is a large solution space of different learning methods and concrete models which can be combined in a stack. Hence, the identification and selection of particular algorithms and instantiations over the time of exploration is crucial for the the user. One way to manage this is to keep track of the history of each model. 
Analysts might also want to step back to a specific previous stage in case they reached a dead end in the exploration of algorithms and models (\textbf{G2}).

\textbf{T3: Manage the performance metrics for enhancing trust in the results.} Many performance or validation metrics are used in the field of ML. For each data set, there might be a different set of metrics to measure the best-performing stacking. Controlling the process by alternating these metrics and observing their influence in the performance can be an advantage (\textbf{G3}).

\textbf{T4: Compare the results of two stages and receive feedback to guide interaction.} To assist the knowledge generation, a comparison between the currently active stack against previously stored versions is important. In general, this includes monitoring the historical process of the stacking ensemble, facilitating interaction and guidance (\textbf{G4}). 

\textbf{T5: Inspect the same view with alternative techniques and visualizations.} To eventually avoid the appearance of cognitive biases, alternative interaction methods and visual representations of the same data from another perspective should be offered to the user (\textbf{G5}). 

%% file: 4.System_Overview.tex
\begin{figure*}[tb]
\centering
\includegraphics[width=\linewidth]{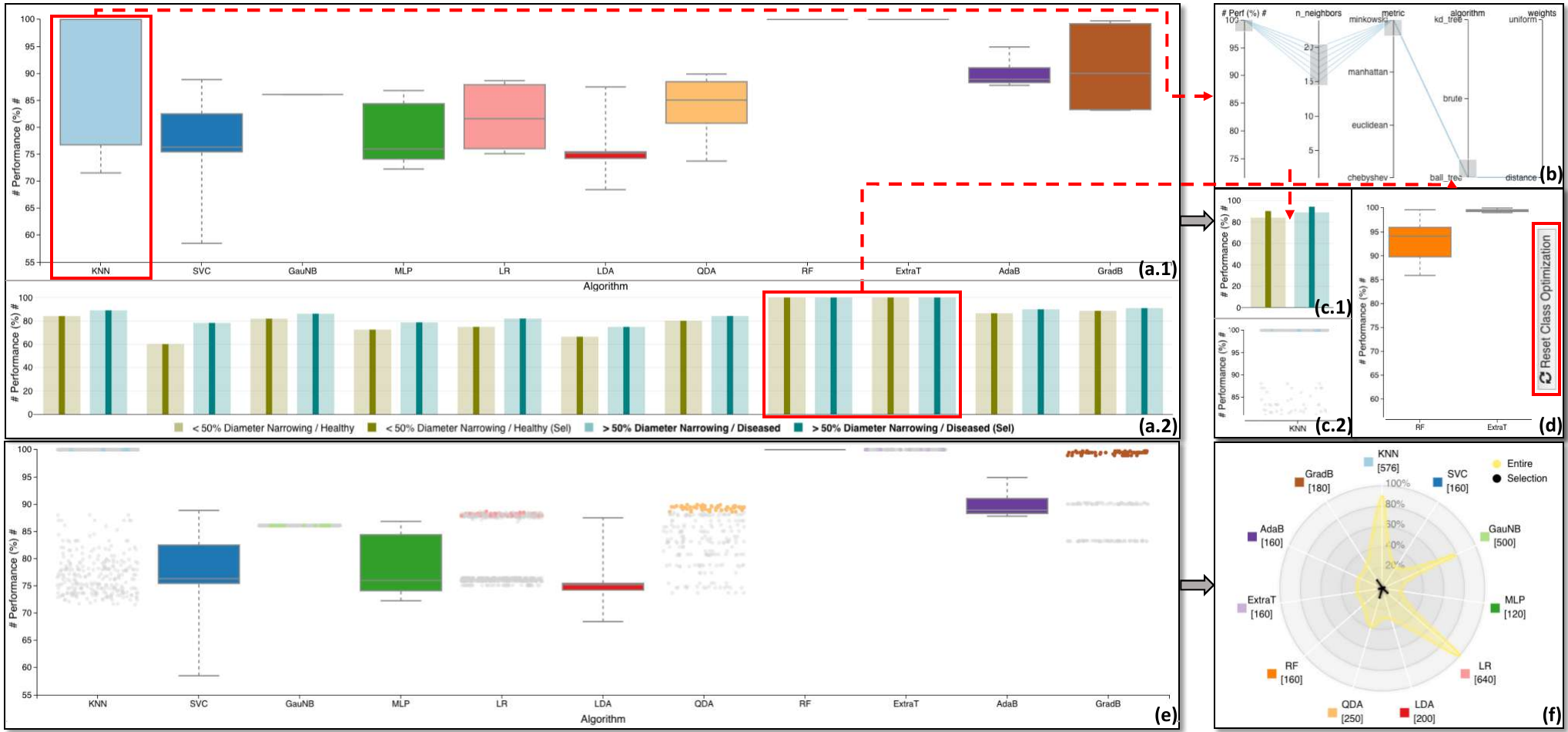}\vspace{-3mm}
\caption{The exploration process of ML algorithms. View (a.1) summarizes the performance of all \emph{available} algorithms, and (a.2) the per-class performance based on precision, recall, and f1-score for each algorithm. (b) presents a selection of parameters for KNN in order to boost the per-class performance shown in (c.1). (c.2) illustrates in light blue the selected models and in gray the remaining ones. Also from (a.2), both RF and ExtraT performances seem to be equal. However in (d), after resetting class optimization, ExtraT models appear to perform better overall. In view (e), the boxplots were replaced by point clouds that represent the individual models of activated algorithms. The color encoding is the same as for the algorithms, but unselected models are greyed out. Finally, the radar chart in (f) displays a portion of the models' space in black that will be used to create the initial stack against the entire exploration space in yellow. The chart axes are normalized from 0 to 100\%.}
\label{fig:use_case1_alg}
\end{figure*}

\begin{figure*}[htb]
  \centering
  \includegraphics[width=\linewidth]{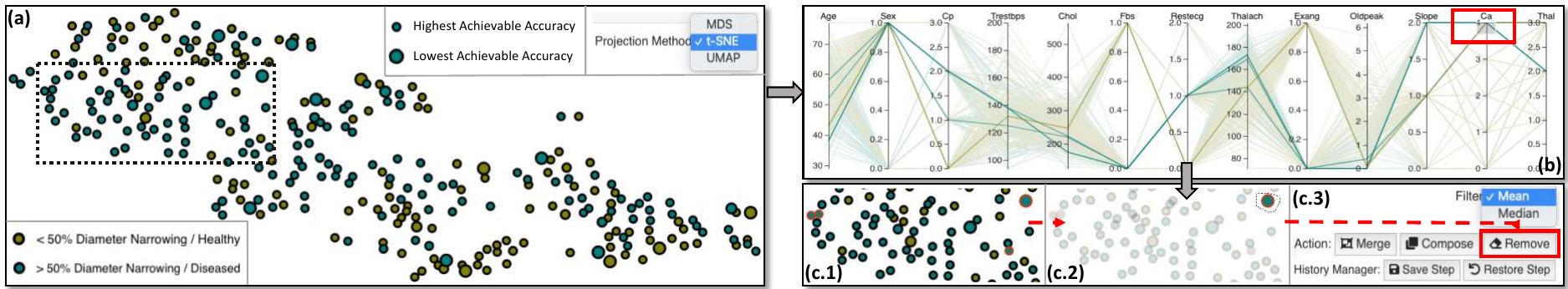}\vspace{-3mm}
  \caption{The data space projection with the importance of each instance measured by the accuracy achieved by the stack models (a). The parallel coordinates plot view for the exploration of the values of the features (b); a problematic case is highlighted in red with values being null (\lq{4\rq} has no meaning for \emph{Ca}). (c.1) shows the brushed instance from the selection in (b) and (c.2) a problematic point that causes troubles to the stacking ensemble. (c.3) indicates the various functionalities that StackGenVis is able to perform for instances.}
  \label{fig:use_case1_ins}
\end{figure*}

Following our design goals and derived analytical tasks, we implemented StackGenVis, an interactive VA system that allows users to build powerful stacking ensembles from scratch. Our system consists of six main interactive visualization panels (see \autoref{fig:teaser}): (1) performance metric selection ($\rightarrow$  \textbf{T3}), (2) history monitoring of stackings ($\rightarrow$  \textbf{T2}), (3) ML algorithm exploration, (4) data wrangling, (5) model exploration ($\rightarrow$  \textbf{T1} and \textbf{T5}), and (6) performance comparison between stacks ($\rightarrow$  \textbf{T4}). We use the following \textbf{workflow} when applying StackGenVis: (i) we choose suitable performance metrics for the data set, which are then used for validation during the entire building process (\autoref{fig:teaser}(a));
(ii) in the next algorithm exploration phase, we compare and choose specific ML algorithms for the ensemble and then proceed with their particular instantiations, i.e., the models;
(iii) during the data wrangling phase, we manipulate the instances and features with two different views for each of them; (iv) model exploration allows us to reduce the size of the stacking ensemble, discard any unnecessary models, and observe the predictions of the models collectively (\autoref{fig:teaser}(d));
and (v) we track the history of the previously stored stacking ensembles in \autoref{fig:teaser}(b) and compare their performances against the \emph{active} stacking ensemble---the one not yet stored in the history---in \autoref{fig:teaser}(c).

StackGenVis works with 11 ML algorithms that can be further subdivided into seven separate groups/types: (a) a neighbor classifier (k-nearest neighbor (KNN)), (b) a support vector machine classifier (SVC), (c) a na{\"\i}ve Bayes classifier (Gaussian (GauNB)), (d) a neural network classifier (multilayer perceptron (MLP)), (e) a linear classifier (logistic regression (LR)), (f) two discriminant analysis classifiers (linear (LDA) and quadratic (QDA)), and (g) four ensemble classifiers (RF, extra trees (ExtraT), AdaB, and GradB).

In the following subsections, we explain the system by using a medicine data set, called \emph{heart disease}, taken from the UCI Machine Learning repository~\cite{Dua2017}. The data set consists of 13 numerical features/attributes and 303 instances. 

\subsection{Data Sets and Performance Metrics} \label{sec:metrics}
As mentioned in \autoref{sec:intro}, the selection of the right performance metrics for different types of analytical problems and/or data sets is challenging.
For example, a medical expert is usually very careful when it comes to handle false negative cases, since human lives may be at stake. In StackGenVis, we offer the option of using eight different metrics with distinct levels of contribution for each, depending on what is appropriate for each individual case.  
The available metrics are grouped into:  threshold ($\rightarrow$ \emph{accuracy}, \emph{g-mean}, \emph{precision}, \emph{recall}, \emph{f-beta score}, and \emph{MCC}); ranking ($\rightarrow$ \emph{ROC AUC}); and probability ($\rightarrow$ \emph{log loss}). 

To illustrate how to choose different metrics (and with which weights), we start our exploration by selecting the \emph{heart disease} data set in \autoref{fig:teaser}(a). Knowing that the data set is balanced, we pick accuracy (weight = 100\%) instead of g-mean (weight = 0\%), as seen in \autoref{fig:teaser}(a). The positive class (\emph{diseased}) is more important than the cases that are healthy, so we use precision and recall instead of ROC AUC (0\%). We also decide that the reproducibility of the results is slightly more important than simply reaching high precision, so we decrease the weight of precision to 80\%. For the f-beta metric, the f2-score is chosen because false negative cases should be better monitored, since they are more important for the underlying problem. MCC is a combination of all f-beta scores and shows us both the false positive and false negative results, which is especially useful for comparing it with the f2-score. Log loss penalizes outliers, and in our case, we should be aware of outliers as we have sensitive healthcare data. Finally, four of the performance metrics include one more option---they are marked with an asterisk in \autoref{fig:teaser}(a)---to compute the individual \textsl{metric} based on \emph{micro-, macro-, or weighted-average}. Micro-average aggregates the contributions of all classes to compute the average metric, whereas macro-average computes the \textsl{metric} independently for each class and then takes the average (therefore treating all classes equally). 
Weighted-average calculates the metrics for each label and finds their average weighted by support (the number of true instances for each label). The data set is a binary classification problem and contains 165 diseased and 138 healthy patients. 
Hence, we choose micro-average to weight the importance of the largest class, even though the impact is low because of the lack of any significant imbalance for the dependent variable. The dice glyphs visible on the right hand side of \autoref{fig:teaser}(a) are static and only used to indicate that specific views do not use all pre-selected metrics. For instance, the performance comparison view \autoref{fig:teaser}(c) only uses four metrics. After this initial tuning of the metrics, we press the \emph{Confirm} button to move further to the exploration of algorithms.

\subsection{Exploration of Algorithms} \label{sec:algs}
\autoref{fig:use_case1_alg}(a.1, a.2) presents the initial views of the 11 algorithms (and their models) currently implemented in StackGenVis.
\autoref{fig:use_case1_alg}(a.1) uses boxplots to represent the performance of the currently unselected algorithms/models based on the metrics combination discussed previously. This compact visual representation provides an overview to users and allows them to decide which algorithms or specific models perform better based on statistical information.
 \autoref{fig:use_case1_alg}(a.2) displays overlapping barcharts for depicting the per-class performances for each algorithm, i.e., two colors for the two classes in our example. The more saturated bar in the center of each class bar represents the altered performance when the parameters of algorithms are modified. Note that the view only supports three performance metrics: precision, recall, and f1-score. 
The y-axes in both figures represent aggregated performance, while the different algorithms are arranged along the x-axis with different colors. 
\autoref{fig:use_case1_alg}(a.1) shows that KNN models perform well, but not all of them. 
We can click the KNN boxplot and further explore and tune the model parameters for KNN with an interactive parallel coordinates plot, as shown in \autoref{fig:use_case1_alg}(b), where six models are selected by filtering.
Wang et al.~\cite{Wang2019ATMSeer} experimented with alternative visualization designs for selecting parameters, and they found that a parallel coordinates plot is a solid representation for this context as it is concise and also not rejected by the users. A drawback is the complexity of it compared to multiple simpler scatterplots.
\autoref{fig:use_case1_alg}(c.1) indicates that, after the parameter tuning, the selected KNN models (narrow, more saturated bars) perform better than the average (wide, less saturated bars) and are thus good picks for our ensemble. Next, we perform similar steps for RF vs. ExtraT without class optimization as shown in \autoref{fig:use_case1_alg}(a.2, d).

Such iterative exploration proceeds for every algorithm until we are satisfied, see \autoref{fig:use_case1_alg}(e) where six algorithms are selected for our initial stack \circled{S1}. 
\autoref{fig:use_case1_alg}(f) shows a radar chart providing an overview of the entire space of available algorithms (yellow contour) against the current selection of models per algorithm (black star plot). 
In brackets, we show the number of all models for each algorithm, together with its name and representative color (\autoref{fig:use_case1_alg}(a.1, e)).

\subsection{Data Wrangling}

\begin{figure*}[tb]
  \centering
  \includegraphics[width=\linewidth]{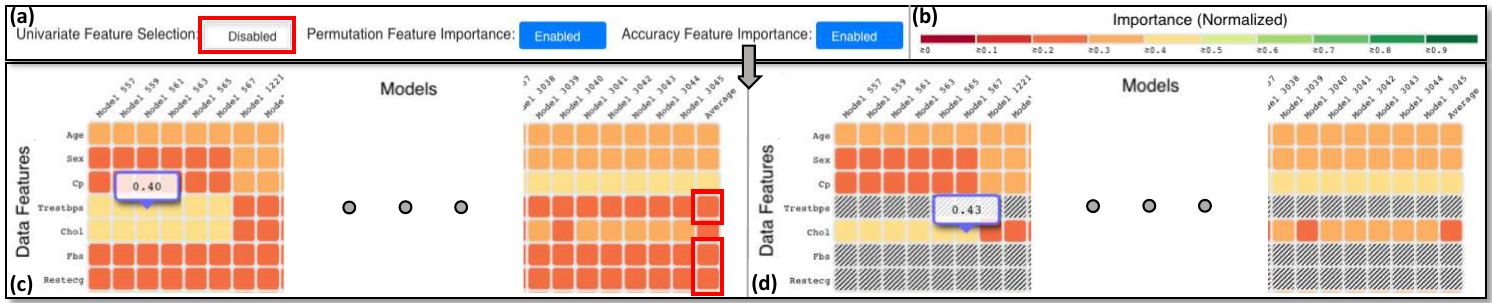}\vspace{-3mm}
  \caption{Our feature selection view that provides three different feature selection techniques. The y-axis of the table heatmap depicts the data set's features,  and the x-axis depicts the selected models in the current stored stack. Univariate-, permutation-, and accuracy-based feature selection is available as long with any combination of them (a). (b) displays the normalized importance color legend. The per-model feature accuracy is depicted in (c), and (d) presents the user's interaction to disable specific features to be used for all the models (only seven features are shown here). This could also happen on an individual basis for every model.}
  \label{fig:use_case1_fea}
\end{figure*}

Pressing the \emph{Execute Stacking Ensemble} button leads to the stacking ensemble shown in \autoref{fig:teaser}(b, \circled{S1}) with the performances shown at the end of the circular barcharts (in \%) and in \autoref{fig:teaser}(c, \circled{S1}). Alternative designs we considered instead of the circular barcharts are standard barcharts or radial plots, but the labels of both would capture more vertical or horizontal space, respectively. In both panels, the performance of the metamodel is monitored with 4 out of the 8 metrics, which are accuracy, precision, recall, and f1-score. The line chart view is linked to the metrics of \autoref{fig:teaser} with a dice glyph showing four. In \autoref{fig:teaser}(c), we encode the active stacking metrics with blue color and the stored stackings of \autoref{fig:teaser}(b, \circled{S1}--\circled{S6}) in red. 

\textbf{Data Space/Data Instances.} 
\autoref{fig:use_case1_ins}(a) is a t-SNE projection~\cite{vanDerMaaten2008Visualizing} of the instances (MDS~\cite{Kruskal1964Multidimensional} and UMAP~\cite{McInnes2018UMAP} are also available in order to empower the users with various perspectives for the same problem, based on the DR guidelines from Schneider et al.~\cite{Schneider2018Integrating}).
The point size is based on the predictive accuracy calculated using all the chosen models, with smaller size encoding higher accuracy value. 
Hence, we want to further investigate cases that cause problems (i.e., we have to look for large points). The parallel coordinates plot in \autoref{fig:use_case1_ins}(b) is used to investigate the features of the data set in detail. 

The \emph{Ca} attribute, for example, has a range of 0--3, but by selection we can see five points with \emph{Ca} values of \lq{4\rq}, see~\autoref{fig:use_case1_ins}(b). These values can be considered as unknown and should be further examined. One of these points belongs to the \emph{healthy} class (due to the olive color) but is very small in \autoref{fig:use_case1_ins}(c.1)---meaning that it does not reduce the accuracy. Four points are part of the \emph{diseased} class. One of those is rather large which affects negatively the prediction accuracy of our classification (see \autoref{fig:use_case1_ins}(c.1) in the upper right corner). In \autoref{fig:use_case1_ins}(c.2), we select the point with our lasso interaction.
We have then several options to manipulate this point as shown in \autoref{fig:use_case1_ins}(c.3): we can remove the point's instance entirely from the data set or merge a set of points into a new one, which receives either their mean or median values per feature.
Similarly, we can compose a new point (i.e., an additional instance) from a set of points. 
The \emph{history manager}  saves the aforementioned manipulations or restores the previous saved step on demand.
For our problematic point, we decide to remove it, and the metamodel performance increases as seen in Step 1 of \autoref{fig:teaser}(c) for the active model in blue. We then store this new stack and get the ensemble shown in \autoref{fig:teaser}(b, \circled{S2}) and \autoref{fig:teaser}(c, \circled{S2}).
The details about  the model's performance and parameters used can also be displayed with a tooltip.

\textbf{Data Features.} For the next stage of the workflow, we focus on the data features. Three different feature selection approaches can be used to compute the importance of each feature for each model in the stack. \emph{Univariate} feature importance is identical for all models, but different for each feature. 
\emph{Permutation} feature importance is measured by observing how random re-shuffling of each predictor influences model performance.
\emph{Accuracy} feature importance removes features one by one, similar to permutation, but then retrains each model by receiving only the accuracy as feedback. These last two approaches are very resource-intensive; therefore, they can be turned off for larger data sets (by disabling \emph{Detailed Feature Search} in \autoref{fig:teaser}(a)). For our example in \autoref{fig:use_case1_fea}(a), they are enabled. 
We normalize the importance from 0 to 1 and use a two-hue color encoding from dark red to dark green to highlight the least to the most important features for our current stored stack, see \autoref{fig:use_case1_fea}(b). The panel in \autoref{fig:use_case1_fea}(c) uses a table heatmap view where data features are mapped to the y-axis (13 attributes, only 7 visible in the figure), and the x-axis represents the selected 204 models of stacking \circled{S2}. The available interactions for this view include panning and zooming in or out. Also, there is a possibility to check the \emph{average value} of all models for each feature, serving as an overview. For our scenario, we can observe that \emph{Trestbps, Chol, Fbs, and Restecg} are less important features. However, \autoref{fig:use_case1_fea}(c, right side) indicates that some models perform slightly better when including the \emph{Chol} feature (due to the less saturated red color). Thus, we only disable the other three attributes by clicking the \emph{Average} buttons in \autoref{fig:use_case1_fea}(c) on the right and get  \autoref{fig:use_case1_fea}(d). After recalculating the performance of the active stacking metamodel (Step 2 of \autoref{fig:teaser}(c)), we store the improved stacking ensemble cf. \autoref{fig:teaser}(b, c, \circled{S3}).

\begin{figure*}[tb]
  \centering
  \includegraphics[width=\linewidth]{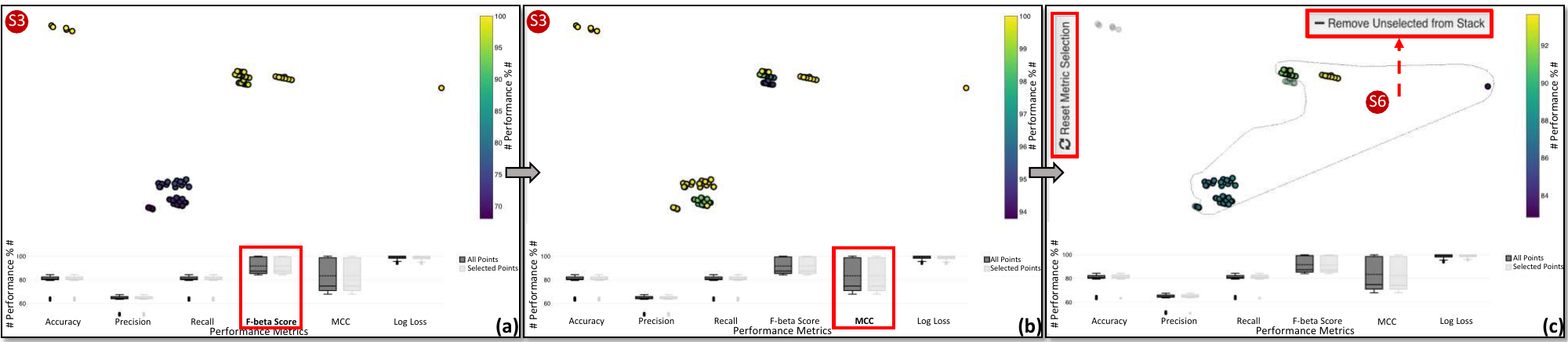}\vspace{-3mm}
  \caption{Visual exploration of the models' space.
  The same MDS projection is observable in varying stages with different legend ranges and diverse colors for each instance, depending on the selected performance metric.
  The three steps in this figure demonstrate that we can reach both performant base models but also diverse algorithms by exploration of different validation metrics in (a) and (b). With the removal of the unselected models in (c), the performance remains stable but the complexity of the stacking ensemble reduces as more models leave the previous stack (cf. \autoref{fig:teaser}(b, \circled{S6})).}
  \label{fig:use_case1_mod}
\end{figure*}

\subsection{Exploration of Models}
The model exploration phase is perhaps the most important step on the way to build a good ensemble. It focuses on comparing and exploring different models both individually and in groups. Due to the page limits, we now assume that we selected the most performant models, removed the remaining from the stack, and reached \circled{S4} (see \autoref{fig:teaser}(b)). Stack \circled{S4} did not boost the performance due to the lack of diverse models from the KNN algorithm (cf. \autoref{fig:teaser}(c)). Diversity is one major component when building stack ensembles from scratch. The performance further drastically fell for \circled{S5} (see \autoref{fig:teaser}(c)) when we reduced the number of models even more (marked as Step 3). As Step 3 led to bad results, we decided to go back to \circled{S3} by clicking the \emph{Stacking Ensemble 3} button in~\autoref{fig:teaser}(b) to reactivate it. 

\textbf{Models' Space.} For the visual exploration of the models shown in \autoref{fig:use_case1_mod}, we use MDS projections (t-SNE or UMAP are also available). 
Each point is one model from the stack, projected from an 8-dimensional space where each dimension of each model is the value of a user-weighted metric. Thus, groups of points represent clusters of models that perform similarly according to all the metrics. 
A summary of the performance of each model according to all selected and user-weighted metrics is color-encoded using the Viridis colormap~\cite{Liu2018Somewhere}. The boxplots below the projection show the performance of the models per metric. 
\autoref{fig:use_case1_mod}(a) presents ensemble \circled{S3}, with all models still included. \autoref{fig:use_case1_mod}(a+b) show the same projection but with different color-encodings for two selected performance metrics: f2-score and MCC. They allow us to decide which models are vital in order to stabilize the performance of the ensemble. For the f2-score (a), the complete cluster of models in dark blue (lower part) does not show good performance results; for MCC (b), the overall performance looks much better except for a small number of models in the center. To get rid of the most underperforming models and keep model diversity at the same time, 
we select, with the lasso tool, the best overall performing models under consideration of the worst performing models for f2-score and MCC (see \autoref{fig:use_case1_mod}(a+b)). We have now a new ensemble \circled{S6} which presents the same results as \circled{S3}, but with 30 fewer models (from 204 to 174 based on six ML algorithms), see Step 4 in \autoref{fig:teaser}(b+c). As such, the complexity of the stacking ensemble has been reduced, and its training can be performed faster without the identified underperforming models. In ~\autoref{fig:teaser}(b, \circled{S6}), we also display the parent stack \circled{S3} from which the final stack has been derived during the workflow. 

\textbf{Predictions' Space.} 
The goal of the predictions' space visualization (\autoref{fig:teaser}(f)) is to show an overview of the performance of all models of the current stack for different instances. 
As in the data space, each point of the projection is an instance of the data set. However, instead of its original features, the instances are characterized as high-dimensional vectors where each dimension represents the prediction of one model. Thus, since there are currently 174 models in \circled{S6}, each instance is a 174-dimensional vector, projected into 2D. Groups of points represent instances that were consistently predicted to be in the same class. In \autoref{fig:teaser}(f), for example, the points in the two clusters in both extremes of the projection (left and right sides, unselected) are well-classified, since they were consistently determined to be in the same class by most models of \circled{S6}. The instances that are in-between these clusters, however, do not have a well-defined profile, since different models classified them differently. After selecting these instances with the lasso tool, the two histograms below the projection in \autoref{fig:teaser}(f) show a comparison of the performance of the available models in the selected points (gray, upside down) vs. all points (black). The x-axis represents the performance according to the user-weighted metrics (in bins of 5\%), and the y-axis shows the number of models in each bin. Our goal here is to look for models in the current stack \circled{S6} that could improve the performance for the selected points. However, by looking at the histograms, it does not look like we can achieve it this time, since all models perform worse in the selected points than in all points.

\subsection{Results of the Metamodel}
Recent work by Latha and Jeeva~\cite{Latha2019Improving} tried out various ensembles for this same data set, such as bagging, boosting, stacking, and majority vote, combined with feature selection. They found that majority vote with the NB, BN, RF, and MLP algorithms was the best combination achieving 85.48\% accuracy. 
For stacking, they reached $\approx$83\% accuracy. 
With StackGenVis, we reached an accuracy of $\approx$88\%, thus surpassing both of their ensembles. 
This shows that our VA approach can be effective when users combine base models to produce the best, most diverse, and simplest possible stacking ensemble.  
The results can be exported in the JSON format (\autoref{fig:teaser}(b, top-right), \emph{Knowledge Extraction}), allowing users to apply the trained stacking ensemble with new data.

%% file: 5.Use_Case.tex
\begin{figure*}[tb]
  \centering
  \includegraphics[width=\linewidth]{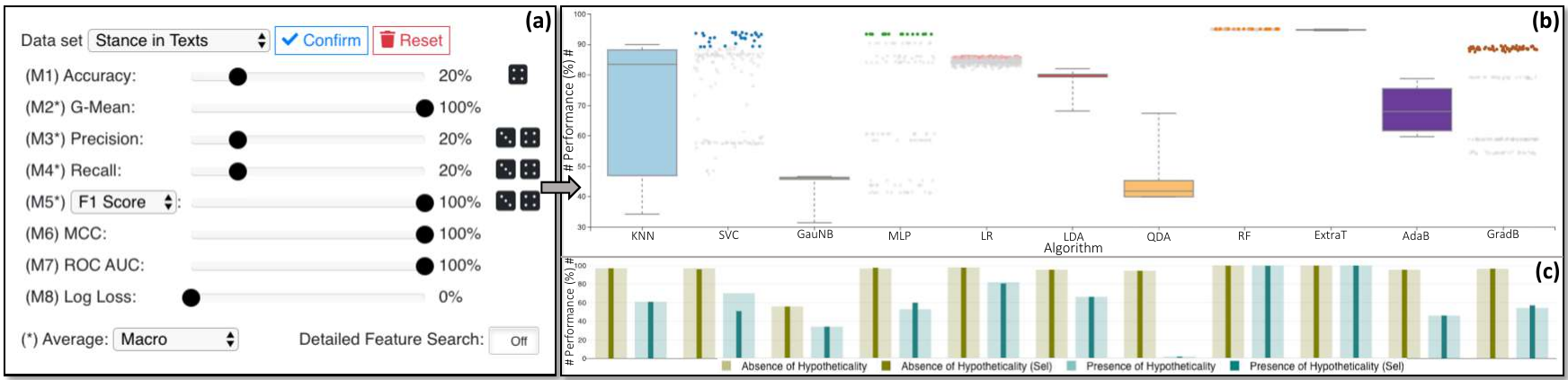}\vspace{-3mm}
  \caption{The process of exploration of distinct algorithms in \emph{hypotheticality} stance analysis. (a) presents the selection of appropriate validation metrics for the specification of the data set. (b) aggregates the information after the exploration of different models and shows the active ones which will be used for the stack in the next step. (c) presents the per-class performance of all the models vs. the active ones per algorithm.}
  \label{fig:use_case2_alg}
\end{figure*}

\begin{figure*}[tb]
  \centering
  \includegraphics[width=\linewidth]{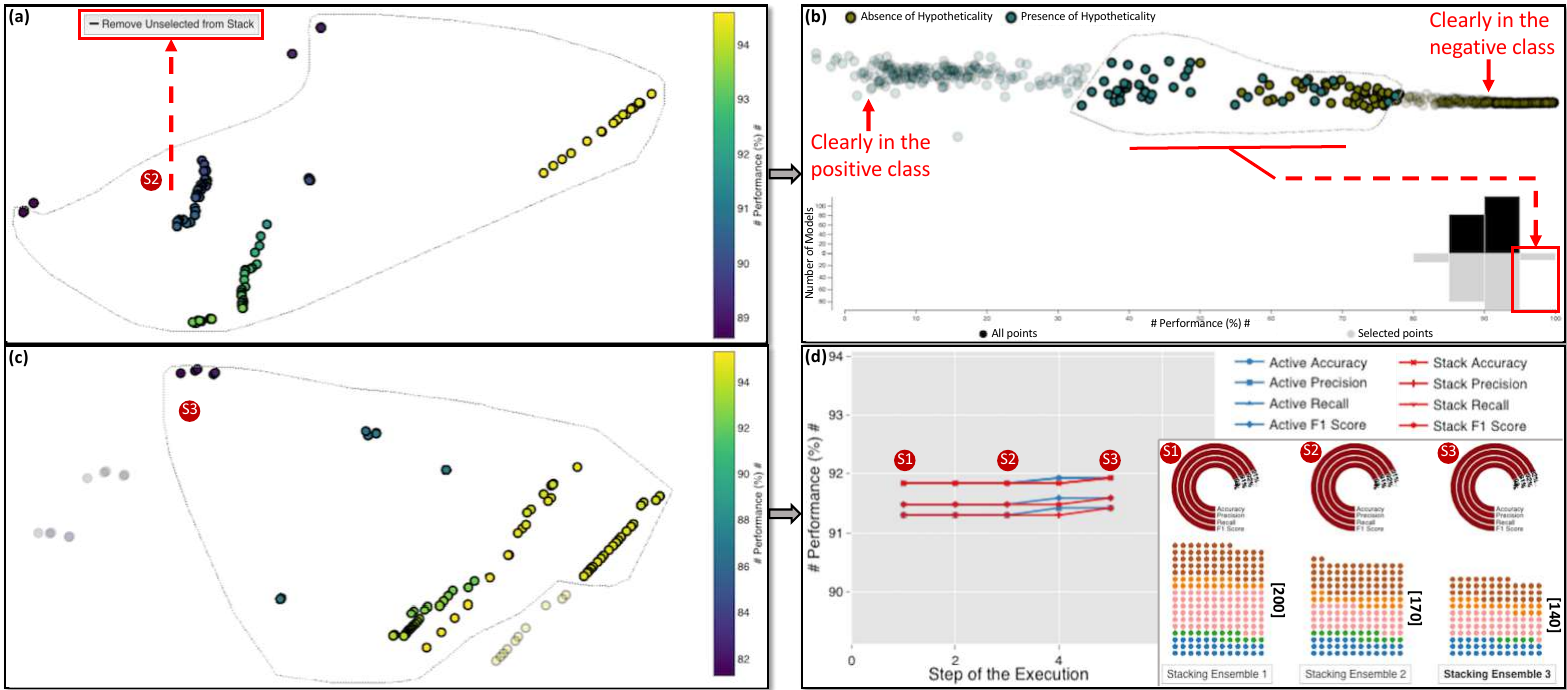}\vspace{-3mm}
  \caption{The exploration of the models' and predictions' spaces and the metamodel's results. (a) presents the initial models' space and how it can be simplified with the removal of unnecessary models. The predictions' space is then updated, and the user is able to select instances that are not well classified by the stack of models in (b). This leads to an updated models' space in (c), where we can even fine-tune and choose diverse concrete models. The results of our actions can always be monitored in the performance line chart and the history preservation for stacks views in (d). 
  }
  \label{fig:use_case2_mod}
\end{figure*}

In this section, we describe how StackGenVis can be used to improve the results of sentiment/stance detection in texts from social media, when compared to previous work from Skeppstedt et al.~\cite{Skeppstedt2017Detection}. The authors studied the automatic detection of seven stance categories: \emph{certainty, uncertainty, hypotheticality, prediction, recommendation, concession/contrast}, and \emph{source}. Their model performed best for the \emph{hypotheticality} category, using a baseline classification approach without the application of heavy feature selection/engineering, therefore we focus on this category in our comparison. It can be considered as a binary classification problem: the \emph{presence} or \emph{absence} of \emph{hypotheticality}.
The training data set was collected using the tool by Kucher et al.~\cite{Kucher2017Active} and it consists of 2,095 instances of annotated training samples. The 300 feature vectors are based on the counts of the most frequent words in the corpus. The data set is very imbalanced, with most cases being on the \emph{absence} side. Skeppstedt et al.~\cite{Skeppstedt2017Detection} used an SVM algorithm to train and build their baseline classifier for this task, and we are going to compare it to our stacking ensemble method in this use case.

\textbf{Selection of Algorithms and Models.} Similar to the workflow described in \autoref{sec:stackgenvis}, we start by setting the most appropriate parameters for the problem (see~\autoref{fig:use_case2_alg}(a)). As the data set is very imbalanced, we emphasize \emph{g-mean} over \emph{accuracy}, and \emph{ROC AUC} over \emph{precision} and \emph{recall}. \emph{Log loss} is disabled because the investigation of outliers is not critical for this text data set, and our computations do not have to be as precise as with medical data. Finally, due to the small number of instances in the presence of the \emph{hypotheticality} class, we start with \emph{macro-average}, which favors the smaller class (the previous work~\cite{Skeppstedt2017Detection} did not explicitly discuss their averaging strategy). The resulting selection of algorithms can be seen in~\autoref{fig:use_case2_alg}(b), where GradB is performing better than AdaB, and RF is slightly better than ExtraT. We improved the per-class performance (as shown in \autoref{fig:use_case2_alg}(c)) by choosing diverse ML models instead of simply the top-performing ones, since LR and RF perform well in the positive class, while other techniques such as SVC and GradB are far better in the negative class. 

\textbf{Optimized Models for Specific Predictions.} In \autoref{fig:use_case2_mod}(a), we see the initial projection of the 200 models selected up to this point (i.e., \circled{S1}). Some models perform well according to our metrics, but others could be removed due to lower performance. However, we should try not to break the balance between performance and diversity of our stacking ensemble.
Thus, we choose to remove some of the models that are positioned close together and are not performing as expected (but not all of such models).
The selection of \circled{S2} leads us to 170 models, cf. \autoref{fig:use_case2_mod}(d). By selecting these models, we get a new prediction space projection, shown in~\autoref{fig:use_case2_mod}(b). While some predictions are clearly in the positive or negative class, we focus on the unclear cases and select them using the lasso tool. The updated histogram indicates in gray that there are better models available for the selected instances. Simultaneously, the models' space is updated as well, depicted in \autoref{fig:use_case2_mod}(c). Again, we try to preserve the diversity, but also reduce the complexity of the ensemble by removing the models with lower performance and low output diversity. As a result, in \autoref{fig:use_case2_mod}(d) we can see that the final stack \circled{S3} contains 140 models that perform better than the previous two stacks of 200 and 170 models. With 5-fold cross-validation, we reach 91\%--92\% performance for all our metamodel's validation metrics. 

\textbf{Evaluation of the Results with the Test Data Set.} To confirm that our findings are solid, we applied the resulting metamodel to the same test data as Skeppstedt et al.~\cite{Skeppstedt2017Detection}, see \autoref{tab:results}. For the \emph{hypotheticality} category, the reported f1-score for the baseline approach was 66\%.
In our case, we reached the following results with the final stack and \emph{weighted-average}: 94.46\% for accuracy, 93.87\% for precision, 94.46\% for recall, and f1-score of 93.87\%. 
Additionally, as an extra validation, we checked the results for the \emph{prediction} category (again as a binary classification problem).
Using our approach, we managed to achieve an f1-score of approximately 82\% compared to 54\% reported by Skeppstedt et al.~\cite{Skeppstedt2017Detection} for the baseline approach.
Finally, it is important to note that, while our approach seems to perform very well for both applications described in this paper, the gain does not come only from the performance. 
Our system supports the exploration and manipulation of many different perspectives of a complex stacking ensemble with the help of visualizations, which is the main burden for stacking ensemble learning to become even more broadly useful. 

\begin{table}[htb]
  \centering
  \caption{Summary of the test data results for stance classification.}
  \label{tab:results}
  \includegraphics[width=\columnwidth]{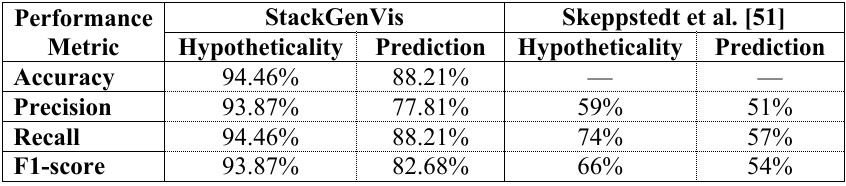}
\end{table}

%% file: 6.Evaluation_Discussion.tex
In this section we discuss the experts' feedback about StackGenVis, as well as possible improvements for our VA approach.

\textbf{Methodology and Information about the Participants.} 
Following guidelines from previous work~\cite{Ma2020Explaining,Ming2020ProtoSteer,Xu2019EnsembleLens}, we conducted semi-structured interviews with three experts to gather qualitative feedback about the effectiveness and usefulness of our system. 
The first expert (\textbf{E1}) is a senior specialist in ML and analytics platforms working in a large multinational company. He has approximately 10 years of experience with ML. 
Moreover, at least half of his PhD studies (2.5 years) was specifically dedicated to stacking ensemble learning. 
The second expert (\textbf{E2}) is a senior researcher in software engineering and applied ML working in a government research institute and as an adjunct professor. He has worked with ML for the past 7 years, and 2 years with stacking ensemble learning. The third expert (\textbf{E3}) is the head of applied ML in a large multinational corporation, working with recommendation systems. She has approximately 7 years of experience with ML, of which 1.5 years are related to stacking ensemble learning. All three experts have a PhD in computer science and none of them reported any colorblindness issues.
The process was as follows: (1) we presented the main goals of our system, (2) we explained the process of improving the \emph{heart disease} data set results (see~\autoref{sec:stackgenvis}), and (3) after that, we gave them a couple of minutes to interact with the VA system by using the simple \emph{iris} data set. 
During this process, we asked them to think aloud, as any feedback might be vital.
However, to structure the process, we explained to them the basic components of our infrastructure that we would like to receive feedback upon. 
Each interview lasted about one hour, during which we recorded the screen and audio for further analysis.
We summarize the key findings from the interviews below.

\textbf{Workflow.} \textbf{E1}, \textbf{E2}, and \textbf{E3} agreed that the workflow of StackGenVis made sense.  
They all suggested that data wrangling could happen before the algorithms' exploration, but also that it is usual to first train a few algorithms and then, based on their predictions, wrangle the data. 
Thus, it is considered an iterative process: the expert might start with the algorithms' exploration and move to the data wrangling, or vice versa. ``The former approach is even more suitable for your VA system, because you use the accuracy of the base ML models as feedback/guidance to the expert in order to understand which instances should be wrangled'', said \textbf{E3}. \textbf{E2} stated that having an evaluation metric from early on is important for benchmarking purposes to choose the best strategy while data scientists and domain experts are collaborating. He also noted that flexibility of the workflow---not forcing the user to use all parts of the VA system for every problem---is an extra benefit.

\textbf{Visualization and Interaction.} \textbf{E1} and \textbf{E3} were positively surprised by the power of visualization regarding the possibilities of dynamically and directly interacting with the ML algorithms and models.
\textbf{E2} added that, after some initial training period (because the system could be a bit overwhelming in the beginning), the power of visualization in StackGenVis for supporting the analytical process is impressive.
\textbf{E3} raised the question: ``why not select the best, or a set of the best models of an algorithm, according to the performance, and why do we need visualization?''
We answered that the per-class performance is also a very important component, and exploratory visualization can assist in the selection process, as seen in~\autoref{fig:use_case1_alg}(b and c.1). 
The expert understood the importance of visualization in that situation, compared to not using it.
Another positive opinion from \textbf{E3} was that, with a few adaptations to the performance metrics, StackGenVis could work with regression or even ranking problems. 
\textbf{E3} also mentioned that supporting \emph{feature generation} in the feature selection phase might be helpful. Finally, \textbf{E1} suggested that the circular barcharts could only show the positive or negative difference compared to the first stored stack. To avoid an asymmetric design and retain a lower complexity level for StackGenVis, we omitted his proposal for the time being, but we consider implementing both methods in the future. 

\textbf{Limitations.} \emph{Efficiency and scalability} were the major concerns raised by \textbf{all the experts}. The inherent computational burden of stacking multiple models still remains, as such complex ensemble learning methods need sufficient resources. Also, the use of VA in between levels makes this even worse.
We believe that, with the rapid development of high-performance hardware and support for parallelism, these challenges are due to diminish in the near future. 
Considering all that, \textbf{E3} noted that our system could be useful in solving competition problems, e.g., on Kaggle, and for her team to run tests before applying specific models to their huge data sets.
Progressive VA workflows~\cite{Stolper2014Progressive} could also be useful for improving the scalability of our approach for larger data sets. 
\emph{Interpretability and explainability} is another challenge (mentioned by \textbf{E3}) in complicated ensemble methods, which is not necessarily always a problem depending on the data and the tasks. However, the utilization of user-selected weights for multiple validation metrics is one way towards interpreting and trusting the results of stacking ensembles. This is an advantage identified by \textbf{E2}. In the first use case we presented to him, he noted that: ``if you are interested in the fairness of the results, you could show with the \emph{history preservation view} of the system how you reached to these predictions without removing the \emph{age} or \emph{sex} features, consequently, not leading to discrimination against patients, for example''.
The visual exploration of stacking methods that use \emph{multiple layers}~\cite{Lorbieski2018Impact} mentioned by \textbf{E1} is set as another future work goal.
While the experts suggested that they almost never continue to stack models into more than one layer in their practice, we can investigate the adaptations for more layers required for our workflow.
Finally, as this work was the first one working with stacking and visualization, we still need to investigate further the impact of \emph{alternative metamodels} on the predictive performance (mentioned by \textbf{E1}) and try out \emph{different modifications of stacking}~\cite{Menahem2009Troika}, for instance, by adapting our workflow with an extra step of visually comparing various metamodels. It is in our plans to conduct a quantitative user study to further evaluate our system in the future.

%% file: 7.Conclusion.tex
In this paper, we introduced an interactive VA system, called StackGenVis, for the alignment of data, algorithms, and models in stacking ensemble learning. The adaptation of an already-existing knowledge generation model leads us to stable design goals and analytical tasks that were realized by StackGenVis. With the careful selection of multiple coordinated views, we allow users to build an effective stacking ensemble from scratch. Exploring the algorithms, the data, and the models from different perspectives and tracking the training process enables users to be sure how to proceed with the development of complex stacks of models that require a combination of not only the best performant but also the most diverse individual models. To retrieve preliminary results about the effectiveness of StackGenVis, we presented use cases with real-world data sets that demonstrated the improvements in performance and the process of achieving them. We also evaluated our approach with expert interviews by retrieving feedback about the workflow of our system, the interactive visualizations, and the limitations of our approach. Those limitations were then identified as future work for further development of our system.